# Self-Supervised Siamese Learning on Stereo Image Pairs for Depth Estimation in Robotic Surgery

M. Ye[1], E. Johns[2], A. Handa[3], L. Zhang[1], P. Pratt[4], G.-Z. Yang[1]

[1]*The Hamlyn Centre for Robotic Surgery, IGHI, Imperial College London, UK*

[2]*Dyson Robotics Laboratory, Imperial College London, UK*

[3]*OpenAI, USA*

[4]*Department of Surgery and Cancer, Imperial College London, UK*
*{menglong.ye, e.johns}@imperial.ac.uk, ankur@openai.com*

## INTRODUCTION

Robotic surgery has become a powerful tool for performing minimally invasive procedures, providing advantages in dexterity, precision, and 3D vision, over traditional surgery. One popular robotic system is the da Vinci surgical platform, which allows preoperative information to be incorporated into live procedures using Augmented Reality (AR). Scene depth estimation is a prerequisite for AR, as accurate registration requires 3D correspondences between preoperative and intraoperative organ models. In the past decade, there has been much progress on depth estimation for surgical scenes, such as using monocular or binocular laparoscopes [1,2]. More recently, advances in deep learning have enabled depth estimation via Convolutional Neural Networks (CNNs) [3], but training requires a large image dataset with ground truth depths. Inspired by [4], we propose a deep learning framework for surgical scene depth estimation using self-supervision for scalable data acquisition. Our framework consists of an autoencoder for depth prediction, and a differentiable spatial transformer for training the autoencoder on stereo image pairs without ground truth depths. Validation was conducted on stereo videos collected in robotic partial nephrectomy.

## MATERIALS AND METHODS

In this work, depth estimation is addressed by training a non-linear function in the form of an autoencoder, outputting per-pixel "inverse depth" (disparity) from a single RGB image. Given stereo image pairs along with intrinsic and extrinsic camera parameters, we formulate a self-supervised deep learning framework as shown in Fig.1. The autoencoder estimates the disparity map $D_l$ from image $\mathbf{I}_l$. This map is then transformed using a Spatial Transformer ST [5], along with image $\mathbf{I}_r$ (counter-part of $\mathbf{I}_l$), to reconstruct $\mathbf{I}_l^*$ via bilinear interpolation. Training the network then requires minimising the reconstruction errors between $\mathbf{I}_l$ and $\mathbf{I}_l^*$.

In this paper, two network architectures for depth estimation are investigated. The first (Fig.1a) is a basic depth estimation network, similar to [4], but modified using, DeConvNet [6] for the autoencoder. For efficient training, we keep the convolutional and deconvolutional layers of DeConvNet, and remove the fully connected layers to reduce the number of parameters in the network. Batch normalisation and rectified linear units are added after each convolutional/deconvolutional layer, and a convolutional layer is included in the last layer to produce the per-pixel disparity map for a total of approximately 31,800,000 parameters (see Table 1). The loss function used for this basic architecture is based on the L1 reconstruction error:

$$l_l = \frac{1}{N}\sum_{i,j}|\mathbf{I}_l(i,j) - \mathbf{I}_l^*(i,j)|. \quad (1)$$

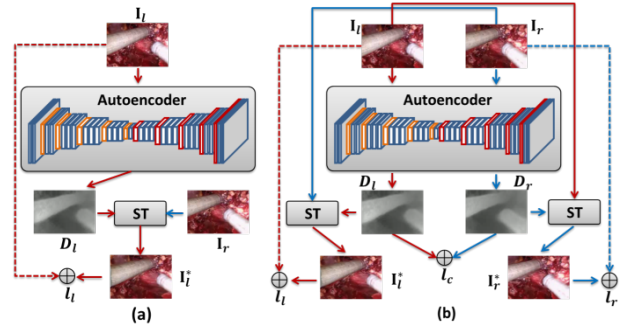

**Fig. 1** Self-supervised depth learning networks. (a) Basic architecture. (b) Siamese architecture.

**Table 1** Autoencoder architecture. Conv: convolution; Deconv: deconvolution; Pool: max pooling; Unpool: max unpooling; In: input channels; Out: output channels; K: kernel size; S: stride.

| Encoder | | Decoder | |
|---|---|---|---|
| Layer | In/Out/K/S | Layer | In/Out/K/S |
| conv1_1 | 3/64/3/1 | unpool5 | -/-/2/2 |
| conv1_2 | 64/64/3/1 | deconv5_1 | 512/512/3/1 |
| pool1 | -/-/2/2 | deconv5_2 | 512/512/3/1 |
| conv2_1 | 64/128/3/1 | deconv5_3 | 512/512/3/1 |
| conv2_2 | 128/128/3/1 | unpool4 | -/-/2/2 |
| pool2 | -/-/2/2 | deconv4_1 | 512/512/3/1 |
| conv3_1 | 128/256/3/1 | deconv4_2 | 512/512/3/1 |
| conv_3_2 | 256/256/3/1 | deconv4_3 | 512/256/3/1 |
| conv_3_3 | 256/256/3/1 | unpool3 | -/-/2/2 |
| pool3 | -/-/2/2 | deconv3_1 | 256/256/3/1 |
| conv4_1 | 256/512/3/1 | deconv3_2 | 256/256/3/1 |
| conv4_2 | 512/512/3/1 | deconv3_3 | 256/128/3/1 |
| conv4_3 | 512/512/3/1 | unpool2 | -/-/2/2 |
| pool4 | -/-/2/2 | deconv2_1 | 128/128/3/1 |
| conv5_1 | 512/512/3/1 | deconv2_2 | 128/64/3/1 |
| conv5_2 | 512/512/3/1 | unpool1 | -/-/2/2 |
| conv5_3 | 512/512/3/1 | deconv1_1 | 64/64/3/1 |
| pool5 | -/-/2/2 | deconv1_2 | 64/3/3/1 |
| conv6 | 512/512/3/1 | conv0 | 3/1/3/1 |

The second architecture is a Siamese network that performs depth estimation on both $\mathbf{I}_l$ and $\mathbf{I}_r$ images (Fig.1b). The two autoencoders have the same structure as the first architecture, but share the same weights during training. This architecture enables us to fully utilise the image dataset, and generalise the network model for both left and right cameras. The loss now is a combination of the outputs from an image pair, and includes the left-right disparity consistency:

$$loss = \alpha_l l_l + \alpha_r l_r + \alpha_c l_c. \qquad (2)$$

Here, $l_r$ is derived similarly as in Eq. 1. $\alpha_l = \alpha_r = 0.5$, and $\alpha_c = 1.0$ weight the individual losses, and $l_c$ is the disparity consistency loss:

$$l_c = \frac{1}{N}\sum_{i,j}|D_l(i,j) - D_r(i+D_l(i,j),j)|. \qquad (3)$$

## RESULTS

Our depth learning networks were implemented using the Torch library on an HP 840 workstation (12 GB NVidia Titan X GPU). We trained both architectures on an *in vivo* dataset collected in a partial nephrectomy procedure performed using a da Vinci Si surgical system. Our dataset includes 20,000 stereo pairs of rectified images, which are randomly sampled over 11 video sequences. The images are resized to 192×96-pixel for efficient training on a GPU.

The Adam optimiser [7] is adopted for loss minimisation. We trained both architectures with 40 epochs on our training dataset. The learning rate was initialised to $10^{-4}$, and reduced by half for every 5 epochs. The batch sizes were 25 (for basic) and 16 (for Siamese), and the total training time were approximately 8 hours and 18 hours, respectively. The trained frameworks only require single image as the input, and take approximately 7ms in depth estimation on a 192×96-pixel image.

For testing, we used a separate video sequence of 3000 stereo pairs of images. We compared our Siamese architecture to the basic architecture, as well as two popular stereo matching approaches, ELAS [8] and SPS [9]. As ground truth depth labels are not available for our *in vivo* surgical data, we evaluate estimated disparity maps using Structural Similarity Index (SSI) [10] (instead of L1 for fair comparisons). For all approaches, SSIs (range of [0.0,1.0]) are calculated between every pair of $\mathbf{I}_l$ and $\mathbf{I}_l^*$, and $\mathbf{I}_r$ and $\mathbf{I}_r^*$ in the testing dataset. Table 2 lists the mean SSI accuracies of all approaches, which shows that the deep learning based approaches outperform the others, with the Siamese architecture achieving the best performance. Finally, example qualitative results of the basic and Siamese networks are presented in Fig. 2, which shows the Siamese network providing more consistent depth estimation than the basic network.

## DISCUSSION

In this work, we have presented self-supervised learning frameworks for depth estimation in surgical images. We introduced a basic depth estimation network which includes an autoencoder and a spatial transformer, and then extended this to a Siamese network which improves the model generalisability. Experiments conducted on *in vivo* videos collected during robotic surgery, showed that our approach performs accurate depth estimation, and outperforms standard stereo matching approaches. Our framework does not require known depth labels during training, and thus provides superior applicability to large-scale *in vivo* video processing where known depths are not available.

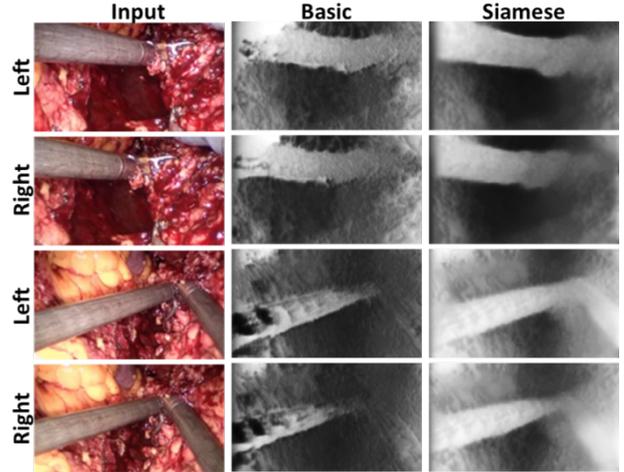

**Fig. 2** Example pairs of image results of the basic and Siamese network architectures. High intensity values indicate close distances to cameras.

**Table 2** SSI measures on image reconstruction quality based on estimated disparity maps. Higher means indicate better performance.

| Methods | ELAS [8] | SPS [9] | Basic | Siamese |
|---|---|---|---|---|
| Mean SSI | 0.473 | 0.547 | 0.555 | **0.604** |
| Std. SSI | 0.079 | 0.092 | 0.106 | **0.066** |